\documentclass[twoside]{article}
\usepackage[accepted]{aistats2018}

\usepackage[comma,authoryear]{natbib}

\newtheorem{theorem}{Theorem}
\newtheorem{lemma}{Lemma}
\newtheorem{proof}{Proof}
\newtheorem{corollary}{Corollary}
\usepackage{graphicx}
\usepackage{color}
\usepackage{bm}
\usepackage{amssymb}
\usepackage{amsmath}
\usepackage{amsfonts}
\usepackage{mathrsfs}
\usepackage{blkarray}
\usepackage{algorithm}
\usepackage{algorithmic}

\makeatletter
\def\@captype{figure}
\makeatother

\begin{document}

%

%

\twocolumn[

\aistatstitle{Efficient Localized Inference for Large Graphical Models}

\aistatsauthor{ Jinglin Chen \And Jian Peng \And Qiang Liu}

\aistatsaddress{ UIUC \And UIUC \And Dartmouth} ]

\begin{abstract}
We propose a new localized inference algorithm for answering marginalization queries in large graphical models with the correlation decay property. Given a query variable and a large graphical model, we define a much smaller model in a local region around the query variable in the target model so that the marginal distribution of the query variable can be accurately approximated. We introduce two approximation error bounds based on the Dobrushin's comparison theorem and apply our bounds to derive a greedy expansion algorithm that efficiently guides the selection of neighbor nodes for  localized inference.  We verify our theoretical bounds on various datasets and demonstrate that our localized inference algorithm can provide fast and accurate approximation for large  graphical models. 
\end{abstract}

\section{Introduction}
Probabilistic graphical models such as Bayesian networks, Markov random fields, and conditional random fields
are powerful tools for modeling complex dependencies over a large number of random variables \citet{koller2009probabilistic,wainwright2008graphical}. 
Graphs are used to represent joint probability distributions, where nodes in a graph denote random variables, and edges represent dependency relationships between different nodes. 
With the specification of a graphical model, a fundamental problem is to calculate the marginal distributions of variables of interest. This problem is closely related to computing the partition function, or the normalization constant of a graphical model, which is known to be intractable and \#P-complete. 
As a result, developing efficient approximation inference algorithms becomes a pressing need. The most popular algorithms include deterministic variational inference and Markov Chain Monte Carlo sampling. 

However,
many challenging practical problems involve very large graphs on which it is computationally expensive to use existing variational inference or Monte Carlo sampling algorithms. This happens, for example, 
when we use Markov random fields to represent the social network of Facebook or use a Bayesian network to model the knowledge graph that is derived from the entire Wikipedia, where in both cases the sizes of the graphical models can be  prohibitively large (e.g., millions or billions of variables). 
It is thus infeasible to perform traditional approximate inference  such as message passing or Mote Carlo on these models because such methods need to traverse the entire model to make an inference. Despite the daunting sizes of large graphical models, in most real-world applications, users only want to make an inference on a set of query variables of interest. The distribution of a query variable is often only dependent on a small number of nearby variables in the graph. As a result, complete inference over the entire graph is not necessary and practical methods should perform inference only with the most relevant variables in local graph regions that are close to the query variables, while
ignoring the variables that are weakly correlated and/or distantly located on the graph. 

In this work, we 
develop a new localized inference method for very large graphical models. Our approach leverages the Dobrushin's comparison theorem that casts explicit bounds based on the correlation decay property in the graphs, in order to restrict the inference to a smaller local region  that is sufficient for the inference of marginal distribution of the query variable. The use of the Dobrushin's comparison theorem allows us to explicitly bound the truncation error which guides the selection of localized region from the original large graph. Extensive experiments demonstrate both the effectiveness of our theoretical bounds and the accuracy of our inference algorithm on a variety of datasets. 

\paragraph{Related  Work}
Approximate inference algorithms of graphical models have been extensively studied in the past decades (see for example \citep{koller2009probabilistic,wainwright2008graphical,dechter2013reasoning} for an overview). 
Query-specific inference, 
including \citep{chechetka2010focused} which proposed a focused belief propagation for query specific inference, and \citep{wick2011query,shi2015learning} which study query-aware sampling algorithms, have recently been introduced for large graphical models. Compared with these methods, our work is theoretically motivated by the Dobrushin's comparison theorem and enables us to efficiently construct the localized region in a principled and practically efficient manner.

\section{Background on Graphical Models}

Graphical models provide a flexible framework for representing relationships between random variables \cite{heinemann2014inferning}. In graph $G$, we use $X=(X_1,X_2,\cdots,X_n)$ to denote a finite collection of $n$ random variables and we use $\bm{x}=(x_1,x_2,\cdots,x_n)$ to refer to an assignment. Suppose $E$ is a set of edges and $\bm{\theta}$ is a set of functions with $\theta_{ij}(x_i,x_j)$ for edge $\langle i\, j\rangle \in E$  and $\theta_i(x_i)$ for node $i \in\{1,2,\cdots,n\}$. We use $\mu(\bm{x};\bm{\theta})$ to represent the joint distribution of the graphical model $(G,E,\bm{\theta})$ as following,
$$\mu(\bm{x};\bm{\theta})=\frac{1}{Z_\mu(\theta)}{\exp\big({\sum_{\langle i\, j\rangle \in E}\theta_{ij}(x_i,x_j)+\sum_i\theta_i(x_i)}\big)},$$
where $Z_\mu(\theta)$ is the normalization constant (also called the partition function).  
In this work, we will focus on the Ising model, an extensively studied graphical model. The Ising model is a pairwise model with binary variables $x_i\in \chi=\{-1,+1\}$. The pairwise and singleton parameters are defined as follows
$$\begin{matrix}
\theta_{ij}(x_i,x_j)=\left(
\begin{array}{cc}
J_{ij} & -J_{ij} \\
-J_{ij} & J_{ij} \\
\end{array}
\right),
& \theta_i(x_i)=\left(
\begin{array}{c}
-h_i    \\
h_i 	\\
\end{array}\right).
\end{matrix}$$
So the distribution of an Ising model is defined as, 
\begin{align}
\label{equ:ising}
\mu(\bm{x};\bm{\theta})=\frac{1}{Z_\mu(\theta)} \exp\big({{\sum_{\langle i\, j\rangle \in E}J_{ij}x_ix_j+\sum_ih_ix_i}\big)}.
\end{align}
Given a graphical model, marginal inference involves calculating the normalization constant, or the marginal probabilities of small subsets of variables. These problems require summation over an exponential number of configurations and are typically \#P-hard in the worst case for loopy graphical models. 
However, practical problems can be often easier than the theoretically worst cases, and it is still possible to obtain efficient approximations by leveraging the special structures of given models. 
In this work, we focus on the query-specific inference, where the goal is to calculate the marginal distribution $\mu(x_i)$ of given individual variable  $x_i$.  For this task, it is possible to make good approximations based on a local region around $x_i$, thus significantly accelerates the inference in very large graphical models. 

\section{Localized Inference and Correlation Decay} 
Given a large graphical model, it is usually not feasible to compute the exact marginal of a specific variable due to the exponential time complexity. Furthermore, it is even not practical to perform the variational approximation algorithms, such as mean field and belief propagation, when the graph is very large. This is because these traditional methods need to traverse the entire graph multiple times before convergence, and thus are prohibitively slow for very large models such as these built on social networks or knowledge bases. 

On the other hand, it is relatively cheap to calculate exact or approximate marginals in small or medium size graphical models. In many applications, users are only interested in certain queries of node marginals. Because users' queries of interest often have strong associations with only a small number of nearby variables in the graph, the complete inference over the full graph is not necessary.  
This can be formally captured by the phenomenon of correlation decay, that is, 
when the graph $G$ is large and sparse, the influence of a random variable on the distribution of another random variable decreases quickly as the distance of the shortest path between the corresponding nodes in the graph $G$ increases. 
The correlation decay property has been widely studied in statistical mechanics and graphical models \citep{rebeschini2015can}.

Formally, assuming that the edge potentials $J_{ij}$ are well bounded, we may expect that variables $x_i$ and $x_j$ are strongly correlated when the distance $d(i,j)$ between node $i$ and $j$ on graph $G$ is small (e.g. $d(i,j)=1,2$), while $x_i$ and $x_j$ may have a rather weak correlation or nearly be independent when node $i$ and node $j$ are far away from each other on $G$. Such property exists broadly in real-world graphical models, such as those built upon social networks in which an individual is mostly influenced by his/her friends. Often, the decaying rate of correlation is negative exponential to the distance $d(i,j)$. 

If a graphical model satisfies the property of correlation decay, it is possible that we can use only the local information in the graph to perform marginal inference, as the distant variables have little correlation with the query variable.
This intuition allows us to use the information from the most relevant variables in the local region close to the queried variable to efficiently approximate its marginal distribution. Assume that $\mu$ is a large graphical model, and we want to calculate a marginal distribution $\mu(x_i)$ of variable $i$. Localized inference constructs a much smaller model $\nu(x_\alpha)$, defined on a small subgraph $\alpha$ that includes $i$, such that $\nu(x_i)\approx \mu(x_i)$. 
The challenge here, however, is how to construct a good localized model and bound its approximation error. We address this problem via the Dobrushin's comparison theorem \citep{follmer1982covariance}, and propose an efficient algorithm to find the local graph region for a given query node and provide an error bound between its approximate and true marginals. 
To get started, we first introduce the Dobrushin's comparison theorem, which is used to compare two Gibbs measures.

\begin{figure}[t]
\begin{center}
\includegraphics[height=4.1cm]{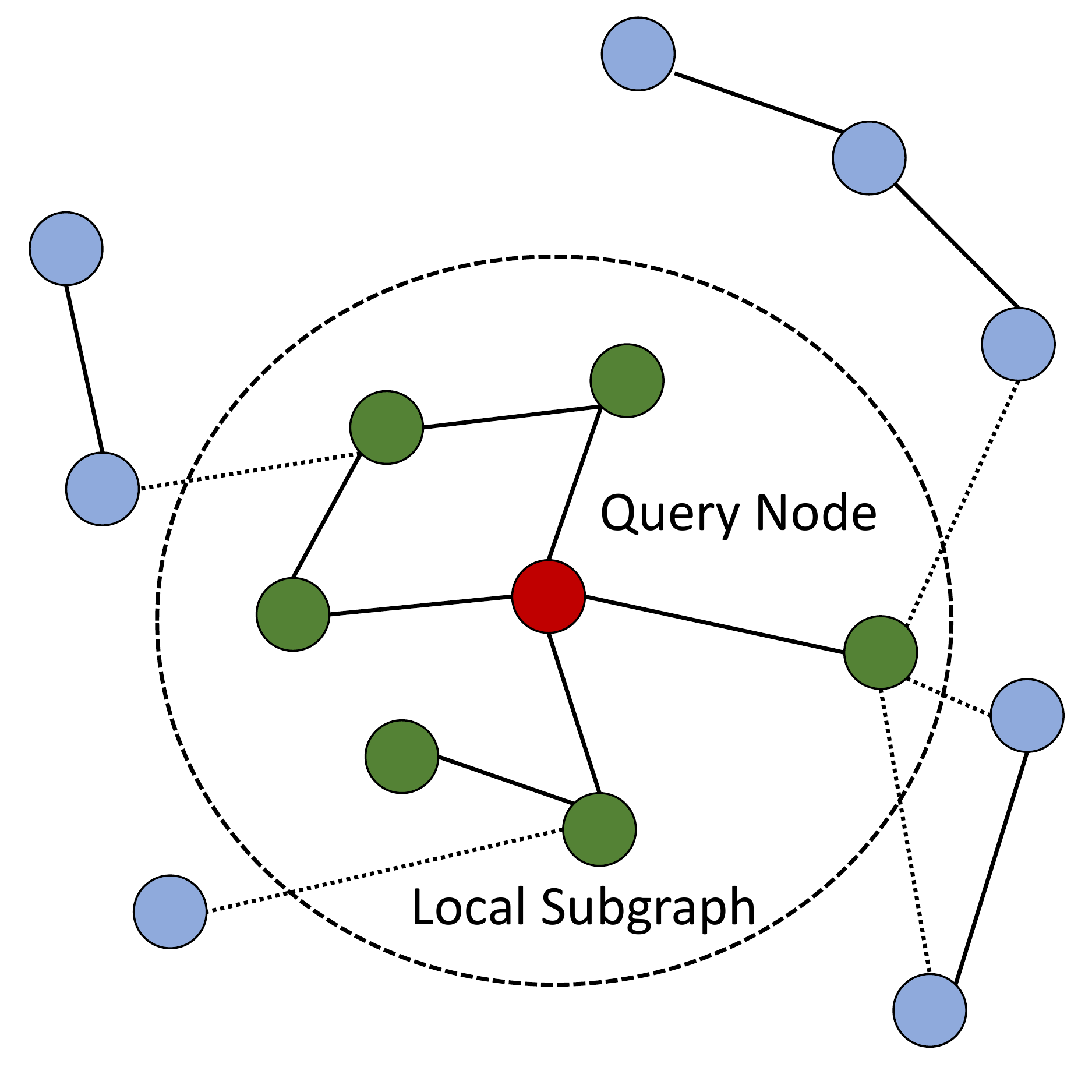}
\caption{The key goal of this work is to approximate queries in large scale graphical models using smaller models on local regions.}
\label{fig4}
\end{center}
\end{figure}

\begin{theorem}{\citep{follmer1982covariance} Dobrushin's comparison theorem }
Let $\mu$ be a Gibbs measure on a finite product space $E=S^I$, where $I$ is an index set. For $i,j\in I$, we define 
$$C_{ij} = \frac{1}{2}\sup\{\|\mu_i(x_i|\bm{x})-\mu_i(y_i|\bm{y})\|:x_k=y_k, \; \forall k\neq j \},$$
where $\mu_i(x_i|\bm{x})$ is the conditional distribution of the $i$th coordinate with respect to the $\sigma$-field generated by the coordinates with index $j\neq i$, and $\|\cdot\|$ is the total variance distance. We compute 
\begin{align}\label{c}
c=\max_{i\in I}\sum_{j\in I} C_{ij},
\end{align}
and assume $c<1$. 
Let $C=(C_{ij})_{i,j\in I}$ and $D=\sum_{n=0}^\infty C^n=(I-C)^{-1}$, then for any probability measure $\nu$ on the same place and any function $f$, we have
\begin{equation*}
\left|\int fd\mu - \int f d\nu\right|\leqslant\sum_{i\in I}(Db)_i\times\delta_i(f),
\label{equ1}
\end{equation*}
where $b=(b_j)$ is the singleton perturbation coefficient of node $x_j$:
\begin{align}\label{equ:b}
b_j=
\frac{1}{2}\sup_{\bm x} \|\mu_j(x_j|\bm{x})-\nu_j(x_j|\bm{x})\|,
\end{align}
and $\delta_i(f)$ is the oscillation of $f$ in the $i$th coordinate, that is, 
$$\delta_i(f) = \max_{x_i,x_i'} |f(\bm x_{\neg i}, x_i) - f(\bm x_{\neg i}, x_i')|.$$
\label{thm1}
\end{theorem}

\begin{figure}[t]
\begin{center}
\includegraphics[height=4.1cm]{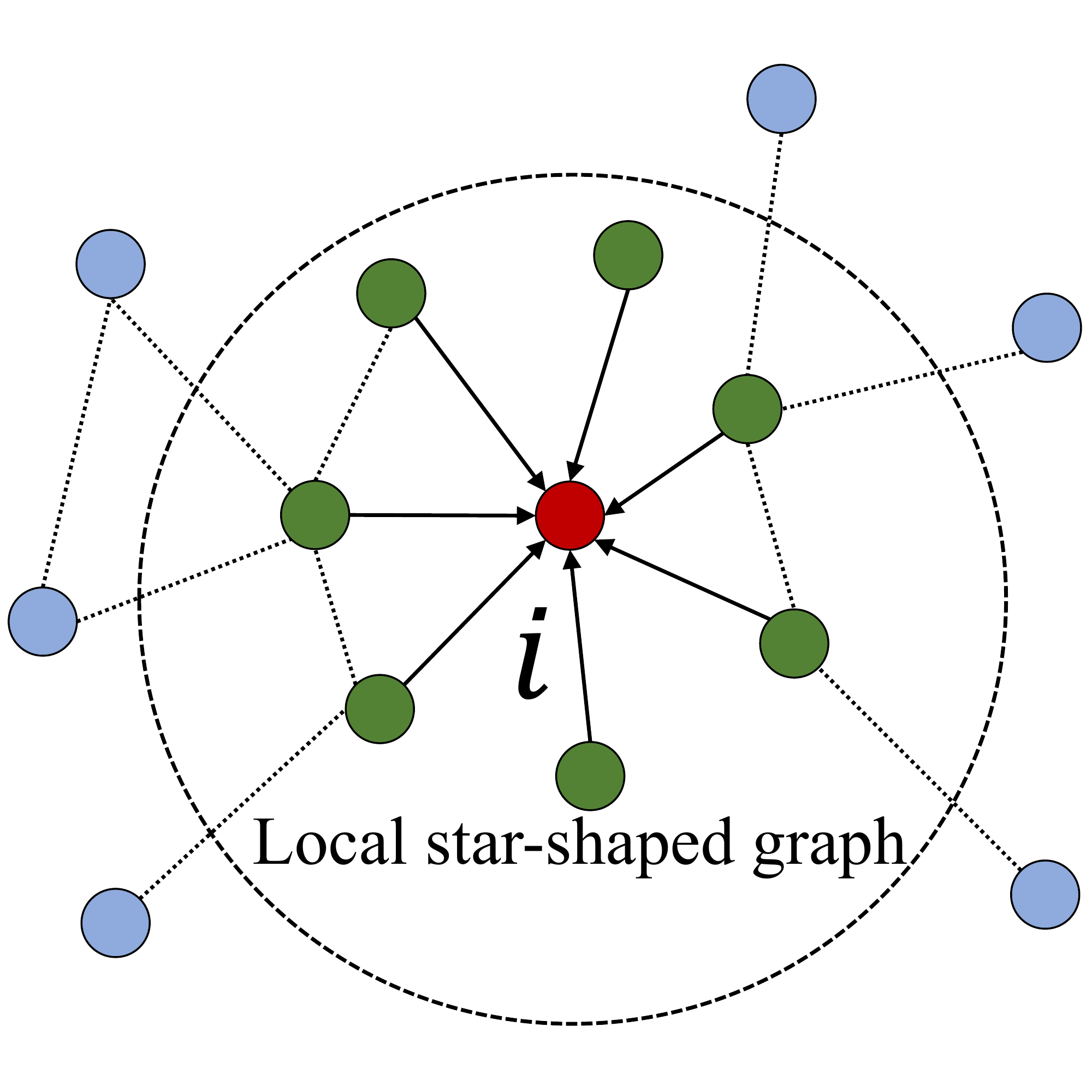}
\caption{Star-shaped graph centered on $i$ consists of nodes within distance 1 to node $i$ and edges represented by solid arrow. In fact, the dashed edges shown in the graph does not exist in the star-shaped graph.}
\label{fig5}
\end{center}
\end{figure}

In Theorem \ref{thm1}, $\mu_i(x_i|\bm{x})$ is the  probability of variable $i$ conditioned on its adjacent variables whose assignments are the same as corresponding entries in $\bm{x}$. According to the Markov property, calculating $\mu_i(x_i|\bm{x})$ only requires information from the local star-shaped graph as shown in Figure \ref{fig5}. It is worth noting that a tighter bound can be obtained by defining $b$ to be 
$b_j = \frac{1}{2}\int \|\mu_j(x_j|\bm{x})-\nu_j(x_j|\bm{x})\|\nu(d\bm{x})$. Here we use the definition in \eqref{equ:b} for lower
computational complexity. The matrix $C$ is known as the Dobrushin's interaction matrix, and the inequality $c=\max_{i\in I}\sum_{j\in I} C_{ij}<1$ is the Dobrushin condition. If this condition holds, the theorem can give us a bound between two measures, which is the result of correlation decay. 

In the following, we will apply Theorem \ref{thm1} to undirected graphical models to derive an approximation bound of marginal distributions. We first denote $I=\{1,2,\cdots,n\}$ by the index set of the variables and assume that we want to query the marginal distribution of variable $x_i$. 
In order to apply Theorem 1, we set $f(\bm{x})$ to be the indicator function of the variable $x_i$, that is, $f(\bm x) = f(x_1,x_2,\cdots,x_n)=\mathbb{I}[x_i=k]$.  Then $|\int fd\mu-\int fd\nu|$ becomes the absolute marginal difference $|\mu(x_i=k)-\nu(x_i=k)|$ between the two measures $\mu$ and $\nu$. In addition, the oscillation of function $f$ is thus reduced to $\delta_i(f)=1$ and $\forall \, j\neq i, \;\delta_j(f)=0$.  
With these simplifications, we obtain a bound of the maximum difference between marginals of the queried node $i$ for two measures: 
\begin{corollary}\label{cor1}
Following the assumptions in Theorem 1 and the above text, we have 
\begin{align}
\max_k|\mu(x_i=k)-\nu(x_i=k)|\leqslant(Db)_i.
\label{equ2}
\end{align}
\end{corollary}
Note that the roles of $\mu$ and $\nu$ in \eqref{equ2} are not symmetric because the Dobrushin coefficient $D$ is solely defined based on $\mu$ (and independent with $\nu$). 
As a result, there are two ways to use bound \eqref{equ2} for localized inference,
depending on whether we treat $\mu$ or $\nu$ as the original model that we want to query or the localized model that we use for approximation, respectively. 
We will next exploit both possibilities in the next sections. 
In Section~\ref{sec:graph}, 
we take $\mu$ as the global model (or measure) and $\nu$ as the localized model (or measure) and derive a simple upper bound relates the approximation error to the distance between the query node and the boundary of the local region on the graph. 
In Section~\ref{sec:greedy}, 
we take $\nu$ as the global model (or measure)
and $\mu$ as the localized model (or measure), we derive another upper bound that only involves the localized region, and leverage it to propose a greedy expansion algorithm to construct the localized model with guaranteed approximations.

\section{Distance-based Upper Bound}
\label{sec:graph}
In this section, we assume that $\mu=\mu(x_1,x_2,\cdots,x_n)$ in Theorem \ref{thm1} is defined by the original graphical model that we want to query, and $\nu$ is a simpler and more tractable distribution that we use to approximate the marginal of $x_i$ in $\mu$.

For notational simplicity, we partition the node-set to two disjoint sets 
$\alpha$ and $\beta =\{1,2,\cdots,n\} \setminus \alpha$, where $\alpha$ is the local subgraph that contains query node $i \in \alpha$ and $\beta$ is the rest of the graph. We use $\partial \alpha$ and $\mathring{\alpha}$ to represent the set of subscripts of nodes on the boundary and in the interior of $\alpha$. Obviously, $\partial \alpha \subseteq \alpha$, $\mathring{\alpha} \subseteq \alpha$, and $\partial \alpha \cup \mathring{\alpha}=\alpha$. Similarly, $\partial \beta \subseteq \beta$, $\mathring{\beta} \subseteq \beta$, and $\partial \beta \cup \mathring{\beta}=\beta$. In addition, we use $x_{\alpha}$ to denote the variables in $\alpha$ and $x_{\beta}$ to denote the variables in $\beta$. 
We will first apply the following lemma to obtain our first result on the relationship between the approximation error of marginals and the radius of the local subgraph $\alpha$.

\begin{lemma}\citep{rebeschini2014comparison}
\label{lem1}
Assume $I$ is a finite set and let $m$ be a pseudo-metric on set $I$. $C=(C_{ij})_{i,j\in I}$ is a non-negative matrix. Suppose that
$$\max_{i\in I}\sum_{j\in I} e^{m(i,j)}C_{ij}\leqslant d < 1.$$
Then matrix $D=\sum\limits_{n = 0}^{\infty}C^n = (I - C)^{-1}$ satisfies
$$\max_{i\in I}\sum_{j \in I} e^{m(i,j)}D_{ij}\leqslant\frac{1}{1-d}.$$
In particular, this implies that
$$\sum_{j\in J}D_{ij}\leqslant\frac{e^{-m(i,J)}}{1-d}$$
for every set $J \subseteq I$, 
where $m(i, J) = \max_{j\in J}m(i,j)$. 
\end{lemma}
This lemma indicates that if $C_{ij}$ decays exponentially with the distance between $i$ and $j$, 
the $D_{ij}$, which is used in Theorem 1 and Corollary 1, also decays exponentially with the distance between $i$ and $j$. 
The condition of this correlation decay lemma is usually mild in practice. When we choose $m\equiv 0$, which is naturally a pseudo-metric, and use Dobrushin's interaction matrix as $C$, the conditions of the lemma hold once the Dobrushin condition is satisfied, because matrix $C$ in Theorem \ref{thm1} is by definition a non-negative matrix and hence $D$ is also non-negative, and every entry in $b$ is less than 1/2.  Applying Lemma \ref{lem1}, we can obtain the following result.
\begin{theorem}
\label{thm2}
Suppose $\mu$ is the probability measure for a graphical model for which we want to query the marginal distribution of node $i$. Let $\nu$ be the another probability on the same space, whose parameters of edges on subgraph $\alpha$ and parameters of nodes in $\mathring{\alpha}$ are the same as $\mu$. Assume the Dobrushin condition holds for $\mu$ ($c=\max_{i\in I}\sum_{j\in I} C_{ij}<1$). 
Let $d(i,\partial \alpha)$ denote the distance between node $i$ and node-set $\partial \alpha$ on the Markov graph $G$ of $\mu$. If we assume
\begin{align}\label{comp}
d(i,\partial\alpha)\geqslant\frac{\ln\frac{t}{2\varepsilon(t-1)(1-c)}}{\ln\frac{1+(t-1)c}{tc}} \text{ for some } t>1,
\end{align}
then $\forall \; \varepsilon > 0$, we have
$$\max_k|\mu(x_i=k)-\nu(x_i=k)|\leqslant \varepsilon.$$
\end{theorem}
 This theorem characterizes the error bound 
 when approximating the global model $\mu$  using another model $\nu$ that matches $\mu$ locally in region $\alpha$. 
Our result shows that in order to 
ensure an $\varepsilon$ bound on the query node $i$,  
the distance $d(i, \partial \alpha)$ from the query node $i$
to the boundary $\partial\alpha$ should be at least linear to $\ln(1/\varepsilon)$.
In other words, the error $\varepsilon$ decreases exponentially with $d(i,\partial \alpha)$. The proof of Theorem 2 can be found in the appendix.

As a result, given $c$ and $\varepsilon$, we can get the minimum value of the lower bound of $d(i,\partial\alpha)$ by optimizing $t$. 
Theorem \ref{thm2} gives a simple but general way to bound the local subgraph of variable $x_i$, as we only need to check the Dobrushin condition and compute $c$ on the whole true graphical model. 

\section{Localized Bound and Greedy Expansion}
\label{sec:greedy} 
The bound in Theorem \ref{thm2} requires computing the value $c$ as defined in \eqref{c} for a given graphical model. However, since $c$ is the maximum $C_{ij}$ of the entire graph, it can be very expensive to compute when the graph is large.  
In this section, we explore another approach 
of using the bound in Corollary~\ref{cor1}, by setting $\nu=\nu(x_1,x_2,\cdots,x_n)$ to be  the distribution of the original graphical model and $\mu$ to be the localized model. 
In this way, we will derive a novel approximation approach by greedily constructing a local graph from the query variable $i$, with guaranteed upper bounds of the approximation error between marginal distributions of $\mu$ and $\nu$. 

To start with, we note that $\nu$ can be decomposed to 
$$\nu(x)=\frac{\psi_\alpha(x_\alpha)\psi_\beta(x_\beta)\psi_{\partial\alpha\partial\beta}(x_{\partial\alpha},x_{\partial\beta})}{Z_\nu},$$
where $\psi_{\alpha}$ is the exponential of a potential function of $x_\alpha$, $\psi_\beta$ is the exponential of potential function of $x_\beta$, and $\psi_{\partial\alpha\partial\beta}$ is the exponential of potential function defined on $x_{\partial\alpha}$ and $x_{\partial\beta}$.

We want to approximate $\nu$ with a simpler model $\mu$ in which 
the nodes in $\alpha$ and $\beta$ are disconnected, so that the inference over $i\in \alpha$ can be performed locally within $\alpha$, irrelevant to the nodes in $\alpha$. 
Formally, we want to approximate $\mu$ by 
$$\mu(x)=\frac{{\psi}_\alpha(x_\alpha){\psi}_\beta(x_\beta)\tilde{\psi}_{\partial\alpha}(x_{\partial\alpha})\tilde{\psi}_{\partial \beta}(x_{\partial\beta})}{Z_\mu},$$
which replaces the  factor $\psi_{\partial \alpha \partial \beta}$ with a product $\tilde \psi_{\partial\alpha}\tilde \psi_{\partial\beta}$ with approximations $\tilde \psi_{\partial \alpha}$ and  $\tilde \psi_{\partial \beta}$.  Therefore, 
the marginal distributions of  $x_{\alpha}$ and $x_{\beta}$ get decoupled in $\mu$, that is,
$$
\mu(x) = \mu(x_\alpha) \mu(x_\beta). 
$$
This decomposition thus allows us to approximately calculate marginal $\mu(x_i)$ efficiently within subgraph $\alpha$. 
The challenges here are 1) how to construct the factors $\tilde\psi_{\partial\alpha}$ and $\tilde\psi_{\partial\beta}$ in $\mu$ to closely approximate $\nu$, 2) how to decide the subgraph region and 3) how to bound the approximation error. We consider two methods for constructing $\tilde\psi$ in this work: 

1. \textbf{[Dropping out]} Simply remove the $\psi_{\partial\alpha\partial\beta}$ in $\nu$. To do so, we set 
\begin{align}\label{equ:cutting}
\tilde\psi_{\partial\alpha}=\tilde\psi_{\partial\beta}=1. 
\end{align}
This corresponds to directly remove all the edges between  $\partial \alpha$ and $\partial \beta$, which is also referred as the ``dropping out'' method in our experiments.
%

2. \textbf{[Mean field]} Find $\tilde \psi_{\partial\alpha}\tilde \psi_{\partial\beta}$  to closely approximate  $\psi_{\partial\alpha\partial\beta}$ by performing a mean field approximation, that is, 
we solve the following optimization problem:  
\begin{align}\label{equ:meanfield}
\min_{\tilde\psi_{\partial\alpha}, \tilde\psi_{\partial\beta}}\mathrm{KL}(\tilde \psi_{\partial\alpha}\tilde \psi_{\partial\beta} ~||~ \psi_{\partial\alpha\partial\beta}),
\end{align}
where the $KL(\cdot ||\cdot)$ refers to the KL divergence of the corresponding normalized distributions. 
To apply the mean field approximation and reduce complexity, we further assume that the nodes are independent in $\tilde{\psi}_{\partial\alpha}$ and $\tilde \psi_{\partial\beta}$. By using the optimized approximation $\tilde{\psi}_{\partial\alpha}$, we will be able to compensate the error of marginal of $x_i$, which is introduced by simply removing the edges between ${\partial\alpha}$ and ${\partial\beta}$, as mentioned in the above. 

Note that the potentials $\tilde\psi_{\partial\beta}$ and $\psi_{\beta}$ in $\mu$ do not influence the calculation of $\mu(x_i)$, for $i\in \alpha$.
For simplicity, we remove all the edges in $\beta$. This will not change the marginal of node $i\in \alpha$. 

By applying Corollary \ref{cor1}, we can now obtain an error bound which, remarkably, only involves the local region $\alpha$. 

\begin{corollary}\label{coro2}
Assume $\mu(x) = \mu(x_\alpha)\mu(x_\beta)$, and the conditions in Theorem 1 holds, we have 
\begin{align}
\max_k|\mu(x_i=k)-\nu(x_i=k)|\leqslant \sum_{j\in\partial\alpha}D_{ij}b_j, 
\label{equ3}
\end{align}
where $b_j$ is defined in Eq~\ref{equ:b}, 
and $D$ is defined by $D_{\alpha\alpha} = (I-C_{\alpha\alpha})^{-1}$; 
here $C$ is defined in Theorem 1. 
\end{corollary}

Note that the upper bound in \eqref{equ3} only involves the local region $\alpha$ and hence can be computed efficiently using mean field or belief propagation within the subgraph on $\alpha$. The proof of Corollary \ref{coro2} and the details on how to calculate $C$ and $D$ for Ising models in practice can be found in the appendix. 

Using the bound in \eqref{equ3},  we propose a greedy algorithm
to expand the local graph starting from query node $i$ incrementally. At iteration, we add a neighboring node that yields the tightest bound using the above bound and repeat this process until the bound is tight enough or a maximum of graph size is reached. This process is summarized in Algorithm~\ref{alg1}.
After we complete the expanding phase, we can apply exact inference or on local region $\alpha$ to calculate the marginal of the query $x_i$ if the size of $\alpha$ is small or perform approximate inference methods if the size of $\alpha$ is medium. The actual size of $\alpha$ can vary in different graphical models, which is mainly determined by the correlation decay property near the query variable $x_i$ or the tightness of the upper bound in Eq (\ref{equ3}).



\begin{algorithm}  
\caption{Greedy expansion algorithm for localized inference} 
\label{alg1}
\begin{algorithmic}[1]
\STATE given a graphical model $\nu$ and a node $i$, approximate marginal probability $\nu(x_i)$
\STATE \textbf{input}: $K=$ maximum number of nodes in the local subgraph $\alpha$ and  $\delta=$ the improvement threshold
\STATE initialize local subgraph $\alpha=\{i\}$ and $bound_{best} = 1$
\WHILE {$|\alpha| < K$ ($|\alpha|$ represents the number of nodes in $\alpha$)}
\STATE set $\beta = \{1,2,\cdots,n\}\setminus \alpha$ and $\partial\beta$ the nodes in $\beta$ that connects with $\alpha$ in $\nu$. 
\FOR {node $k\in\partial\beta$}
\STATE add node $k$ to $\alpha$ and get a candidate local subgraph  $\alpha_k^{new}=\alpha \cup \{k\}$. 
\STATE construct local model $\mu$ by setting $\tilde{\psi}_{\partial\alpha}=1$ (dropping out, Eq~\eqref{equ:cutting}) or estimating it using mean field as in Eq~\eqref{equ:meanfield}.  
\STATE calculate the bound $bound_k$ in \eqref{equ3} where the $bound_k$ refers to $\sum_{j\in\partial\alpha_k^{new}}D_{ij}b_j$
\ENDFOR
\IF{$\min_{k\in \partial\beta} bound_k<bound_{best}-\delta$}
\STATE update $bound_{best}=\min_{k\in \partial\beta} bound_k$
\STATE update $\alpha=\alpha\cup \{\mathop{\mathrm{argmin}}_{k\in\partial\beta} bound_k\}$
\ENDIF
\ENDWHILE
\end{algorithmic}  
\end{algorithm}


\subsection*{Computational Complexity}
Here we consider the computational complexity of expanding the local subgraph and the complexity of localized inference. We always suppose that the maximum degree of the graph $G$ is $d$ and we define the maximum distance between the query node and any node in the subgraph to be the radius of the subgraph. 

First, given a threshold $\varepsilon$, from Theorem \ref{thm2}, we just need a subgraph with radius 
$$r = \inf_{t > 1}\ \left\lceil\frac{\ln\frac{t}{2\varepsilon(t-1)(1-c)}}{\ln\frac{1+(t-1)c}{tc}}\right\rceil,  
$$
where we recall that $c$ is the Dobrushin coefficient $c = \max_{i\in I}\sum_{j\in I} C_{ij}$. 
In particular, taking $t=2$ shows that we just need $r = \lceil -\ln(\epsilon(1-c))/\ln\frac{1+c}{2c} \rceil $.
It is worth noting that $r$ decreases when $c$ becomes small and/or the accuracy threshold $\varepsilon$ becomes large. 
Since the size of the subgraph with radius $r$ is no more than 
$1+d+d^2+\cdots+d^{r}=(d^{r+1}-1)/(d-1)$, it can be much smaller than the whole graph. 
As a result, the inference over the subgraph is much more efficient. 

Then, we discuss the computation complexity in each expansion step (Algorithm~1, line 6-10). We need to loop over the nodes in $\partial\beta$. In the loop, we need to calculate the vector $b$ and the matrix $D$. The calculation for each element in $b$ requires the enumeration of different assignments in the neighborhood of such node, which is bounded because it is not related to the size of the whole graph. In the calculation of matrix $C$, we only need to update the elements related to the new node. The number of such elements is no more than $d$ and the calculation of each element is not related to the size of the whole graph. $D$ can be derived from $C$ and use historical information to calculate incrementally. The complexity is no more than computing the inverse of the whole matrix $I-C_{\alpha\alpha}$. If we use mean field approximation in the greedy expansion, the computation is also cheap because the sizes of $\partial\alpha$ and $\partial\beta$ are small.

\section{Experiments}
We test our algorithm on both simulated and real-world datasets. The results indicate that our method provides an efficient localized inference technique.  

\subsection{2D Ising Grid}
In this section, we perform experiments on 2D-grid Ising models and regard the localized probability as $\nu$ and regard the true probability as $\mu$. 
The graph is a $10 \times 10$ lattice and the coordinate of query node is $(5,5)$. 
The parameters in the Ising model is generated by drawing $h_i$  uniformly from $[-I_1,I_1]$ for all nodes $i$ and $J_{ij}$ uniformly from $[-I_2,I_2]$ for all edges $\langle i\, j\rangle$. 
Here $I_1$ and $I_2$ control the locality and hardness of this Ising model.



\begin{figure}[tb]
\begin{center}
\includegraphics[height=6.5cm]{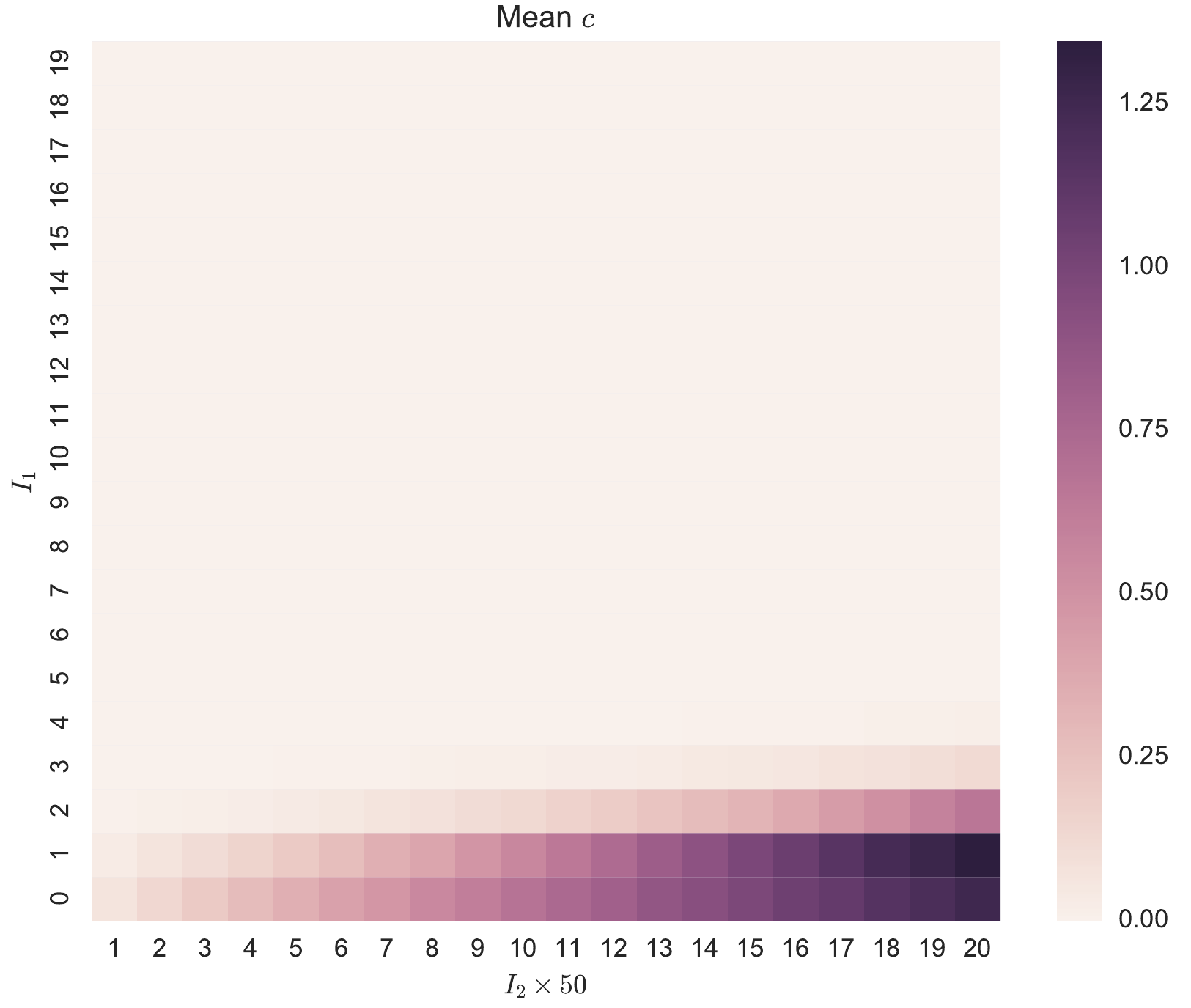}
\caption{Verifying Dobrushin's condition. The color denotes the value $c$ in \eqref{c} of models generated with different values of $I_1$ and $I_2$.}
\label{fig11}
\end{center}
\vspace{-0.4cm}
\end{figure}

\textbf{Checking Dobrushin's condition}
We start with numerically checking the Dobrushin condition $c = \max_i\sum_jC_{ij}<1$. 
In Figure~\ref{fig11}, we show the values of $c$ for Ising models generated with different values of $I_1$ and $I_2$, using a heatmap. We can see that $c$ is smaller than one in most regions, but is larger than one when $I_2$ is very large and $I_1$ is very small, in which case the nodes are strongly coupled together (no correlation decay) and there is no significant local information. 
The hope, however, is that real problems tend to be easier because a large amount of information is available. 

\textbf{Comparing Different Expansion Algorithms}
In this part, we compare the true approximation error $\max_{k}|\mu(x_i=k) - \nu(x_i=k)|$ to the bound $\sum_{j\in\partial\alpha}D_{ij}b_j$ given by our algorithm when we expand the local subgraph. 
The true error $\max_{k}|\mu(x_i=k) - \nu(x_i=k)|$ is evaluated using the brute-force algorithm. 
When removing the bipartite graph, we try both simply dropping edges and the mean field approximation.
In all the experiments, we use the UGM Matlab package\footnote{http://www.cs.ubc.ca/~schmidtm/Software/UGM.html} 
for the mean field approximation. 

In order to better compare the error, we also add two baselines. One baseline is that we expand the local subgraph in each step by randomly selecting a node in the boundary $\partial \beta$. Another baseline is that we expand the local subgraph greedily by choosing the node in $\partial \beta$ that has the maximum $L_2$ norm over the edge-set between such node and the subgraph $\alpha$. Formally, 
consider the Ising model in \eqref{equ:ising} whose weight on each edge is $J_{ij}$,  
the nodes we we add in each expansion should be 
$
i^*=\arg\max_{i\in \partial \beta}\sum_{j\in\alpha} J_{ij}^2. 
$
The intuition is that when 
the magnitude of the edges weights is large, the node may be more related to the nodes in the subgraph.

\begin{figure}[tb]
\begin{center}
\includegraphics[height=5.5cm]{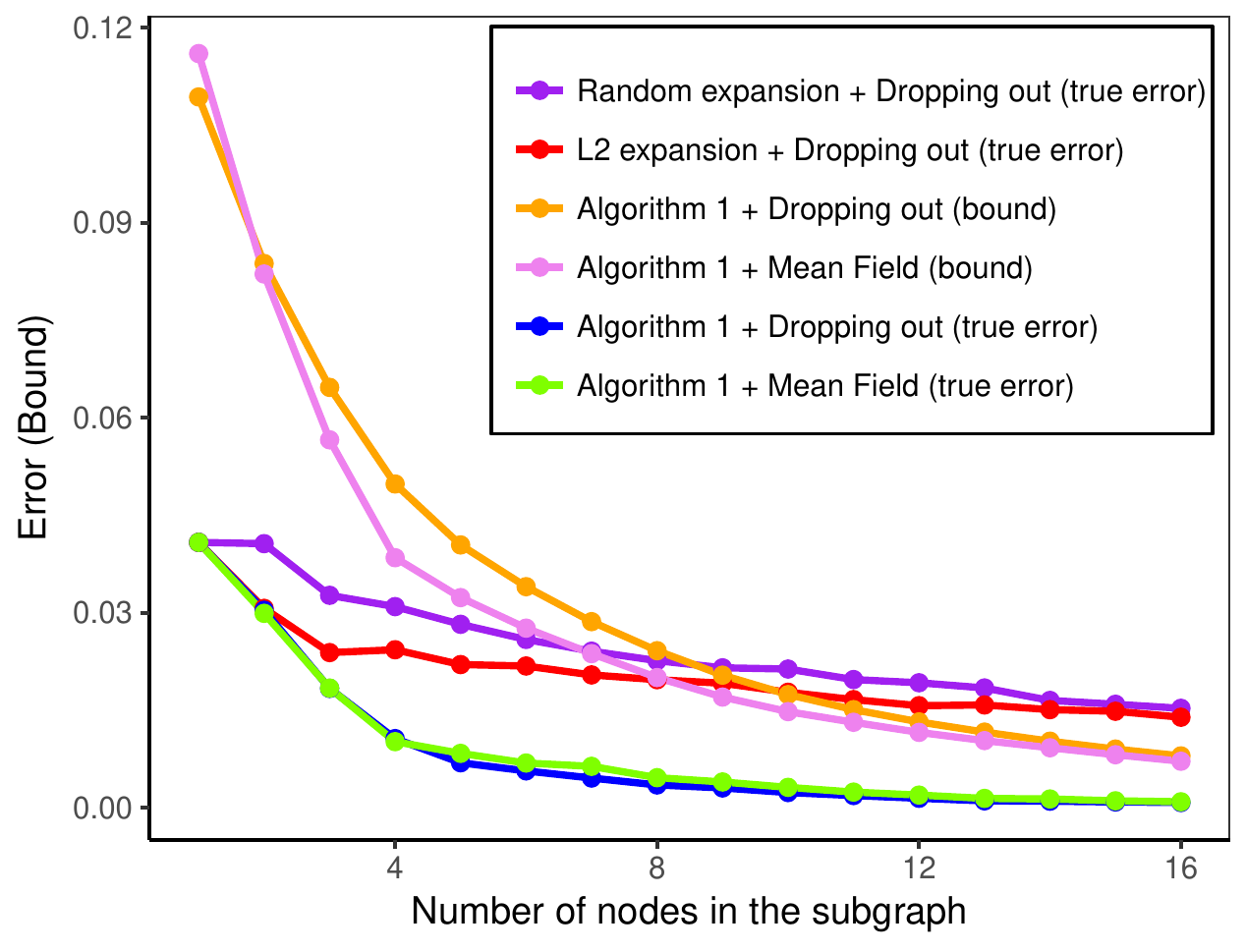}
\caption{The true errors and our upper bounds for the marginal approximation when we expand the local subgraphs to different sizes.}
\label{fig2}
\end{center}
\end{figure}

In Figure~\ref{fig2}, we compare our greedy expansion method stated in Algorithm \ref{alg1} and the baselines stated above to construct the local graph incrementally. For this experiment, 
we fix $I_1=1$ and $I_2=0.25$ and average on 100 random trials. 
We stop expanding the graph when the local subgraph contains 16 nodes. We calculate the mean value of the true errors and bounds in the 100 trials for a different number of nodes in the subgraph. 

From Figure~\ref{fig2}, we can seer that, when combined the dropping out method for constructing $\tilde \psi$, 
our greedy expansion method significantly outperforms the two baselines. 
We also find that the mean field method for constructing $\tilde\psi$ 
gives about the same true error as the dropping out method, but provides a tighter upper bound. 
It is interesting to note that 
the true errors of the two baseline expansion methods are sometimes even worse than the upper bounds of our greedy expansion, indicating the strong advantage of our method. 

\begin{figure}[tb]
\begin{center}
\includegraphics[height=5.5cm]{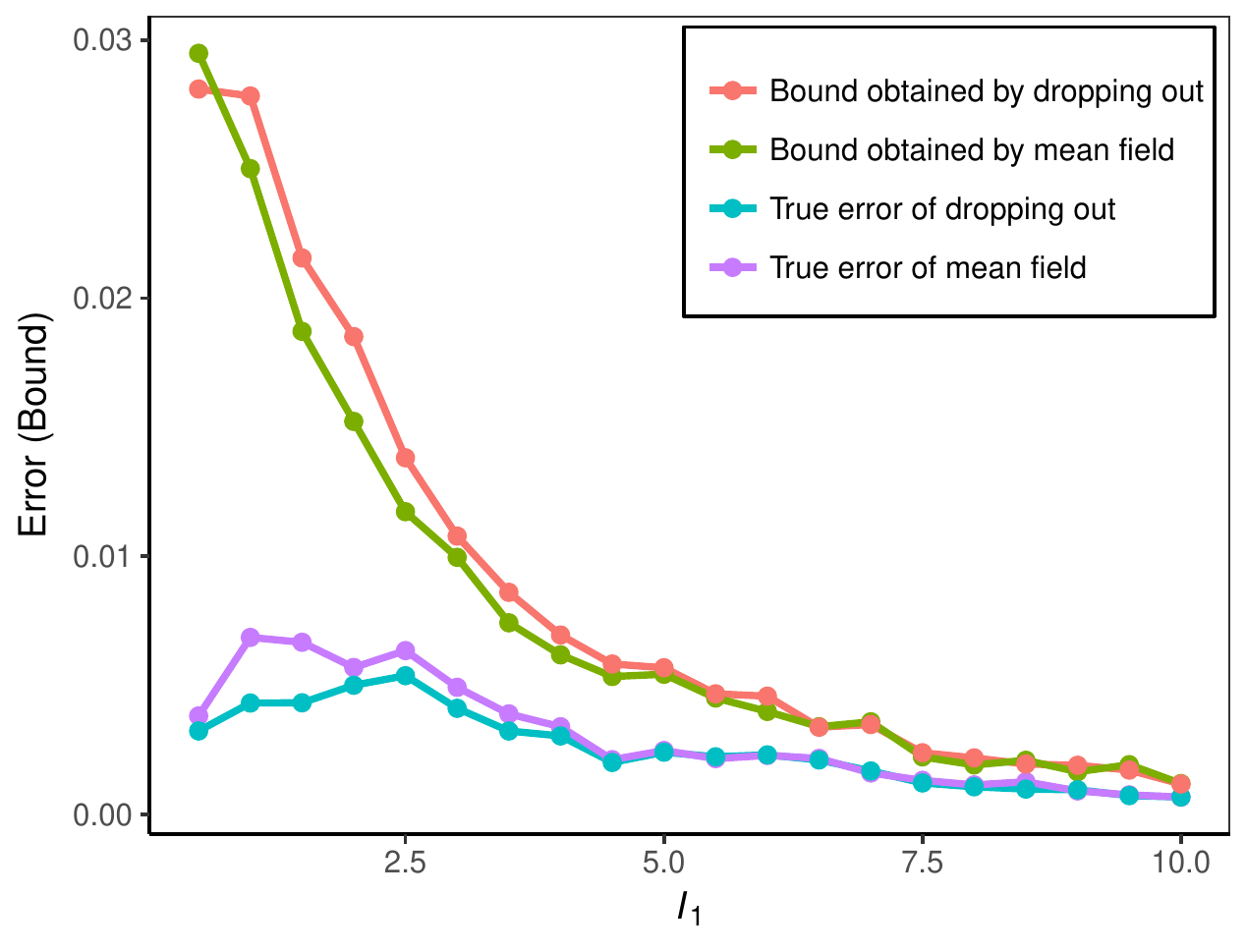}
\caption{Our bounds and the true errors vs. different $I_1$.}
\label{fig9}
\end{center}
\vspace{-0.5cm}
\end{figure}

We further investigate 
how the parameters of the Ising model may influence the results of the algorithms and the tightness of the bound. 
For this purpose, 
we fix $I_2=0.25$ and vary $I_1$ 
in the range of $\{0,0.5,1,\cdots,10
\}$ 
in Figure~\ref{fig9}. 
For each setting, we simulate 100 times and then calculate the mean error and bound. 
From Figure \ref{fig9}, we can find that the bound is again relatively tight, especially 
when the value of $I_1$ is large. both the bounds and the true errors decrease as $I_1$ increases because the  correlation  decay is stronger and the inference task is easier with strong local evidence on the singleton potentials (large $I_1$). 

\subsection{Cora data set}
We perform experimental evaluations on the Cora data set\footnote{https://people.cs.umass.edu/~mccallum/data.html}. 
Cora consists of a large collection of machine learning papers with citation relations between the papers, in which each paper is labeled as one of seven classes. 
For our experiment, we binarize the labels by taking 
``Neural Networks'' as label 1 and the remaining classes as label $-1$. 
We process the data by removing the hubs in the graph and truncate the graph to have a maximum degree of 15; this is done by randomly deleting edges of the nodes whose degree is larger than 15 until the whole graph is degree bounded by 15. We then experiment on the maximum connected subgraph, which consists of 2389 nodes and 4325 edges.

In order to construct an Ising model based on Cora,
we random draw edge potentials by $J_{ij} \sim N(0.25,0.05)$ for each edge of the citation graph, and draw the singleton 
potentials by $h_i \sim N(0.1I_1,1)$ for  nodes with true label $1$,
and $h_i \sim N(-0.1I_1,1)$ for nodes with true label $-1$.
Here $I_1$ is a parameter that we choose from $\{0,1,\cdots,10\}$. 
When $I_1$ increases from 0 to 10, the node potentials increases so that marginal is more dominated by the status of the query node and the querying is more easily. 

\textbf{Comparing local inference with global inference}
In this part, we want to compare the performance of inference on the local graph to the inference on the global graph. Since the global graph is too large, we can only use approximate inference algorithm. Here, we use mean field to do the global inference and use it as a baseline. For the local graph, we  expand the graph greedily as stated in Algorithm \ref{alg1} and  choose a threshold of $\delta=0.005$ and stop expanding when the subgraph already has 16 nodes. 

\begin{figure}[H]
\begin{center}
\includegraphics[height=5.4cm]{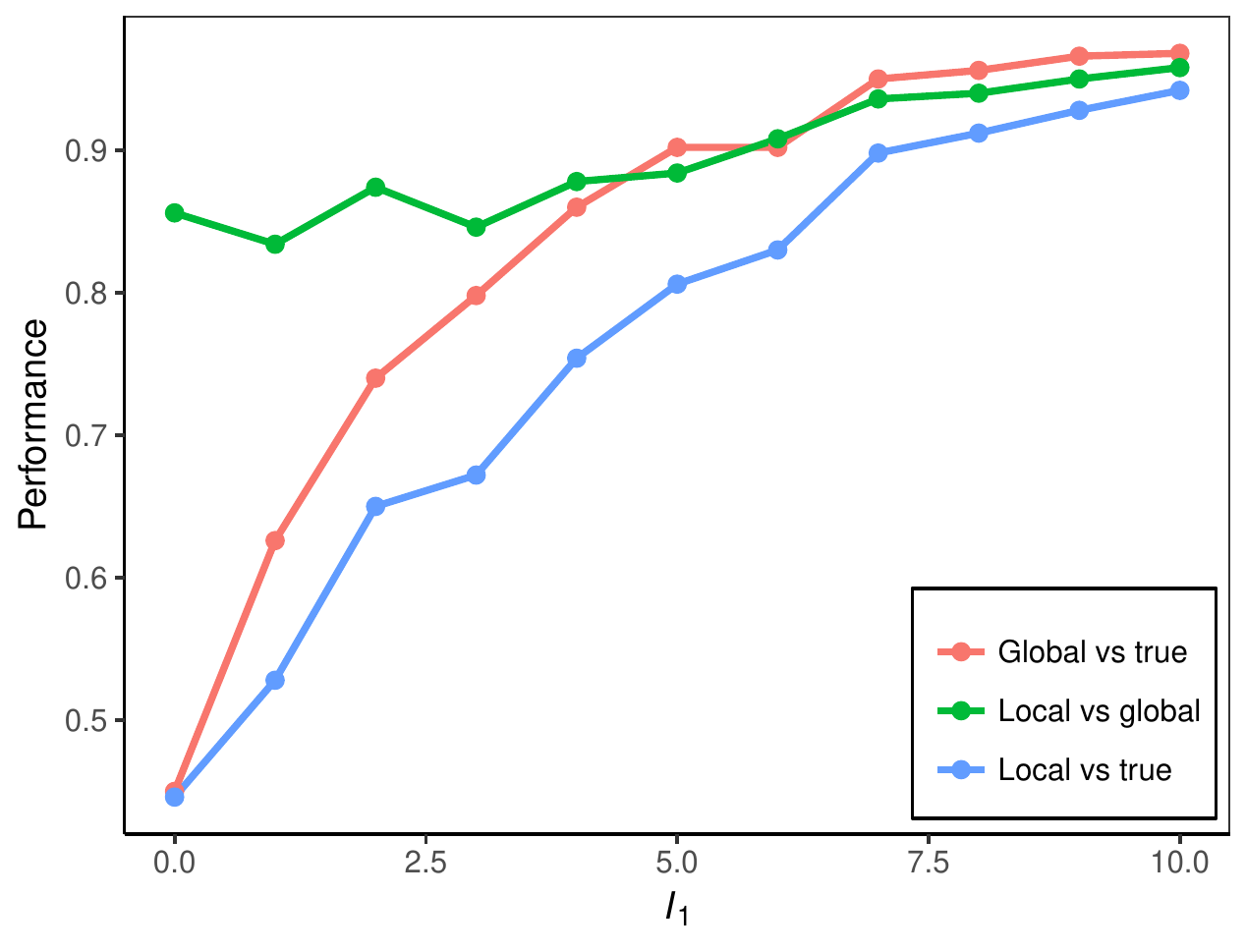}
\caption{The accuracy of different algorithms when $I_1$ changes. Red and blue: the accuracy of the labels given by the global and the local inference evaluated w.r.t. the true labels. Green: the accuracy of the local inference evaluated w.r.t. the labels provided by the global inference.}
\label{fig8}
\vspace{-0.3cm}
\end{center}
\end{figure} \vspace{-0.1cm}
For $I_1 \in \{0,1,\cdots,10\}$, we query the same 500 nodes randomly selected out of the 2389 nodes and evaluate their marginal distributions. In global inference and local inference, we have the marginal on the each query node. 
If the marginal is larger than 0.5, 
we consider our inference algorithm give it label 1, whereas if the marginal is less than 0.5, we 
give it label $-1$. 

\begin{figure}[tb]
\begin{center}
\includegraphics[height=5.4cm]{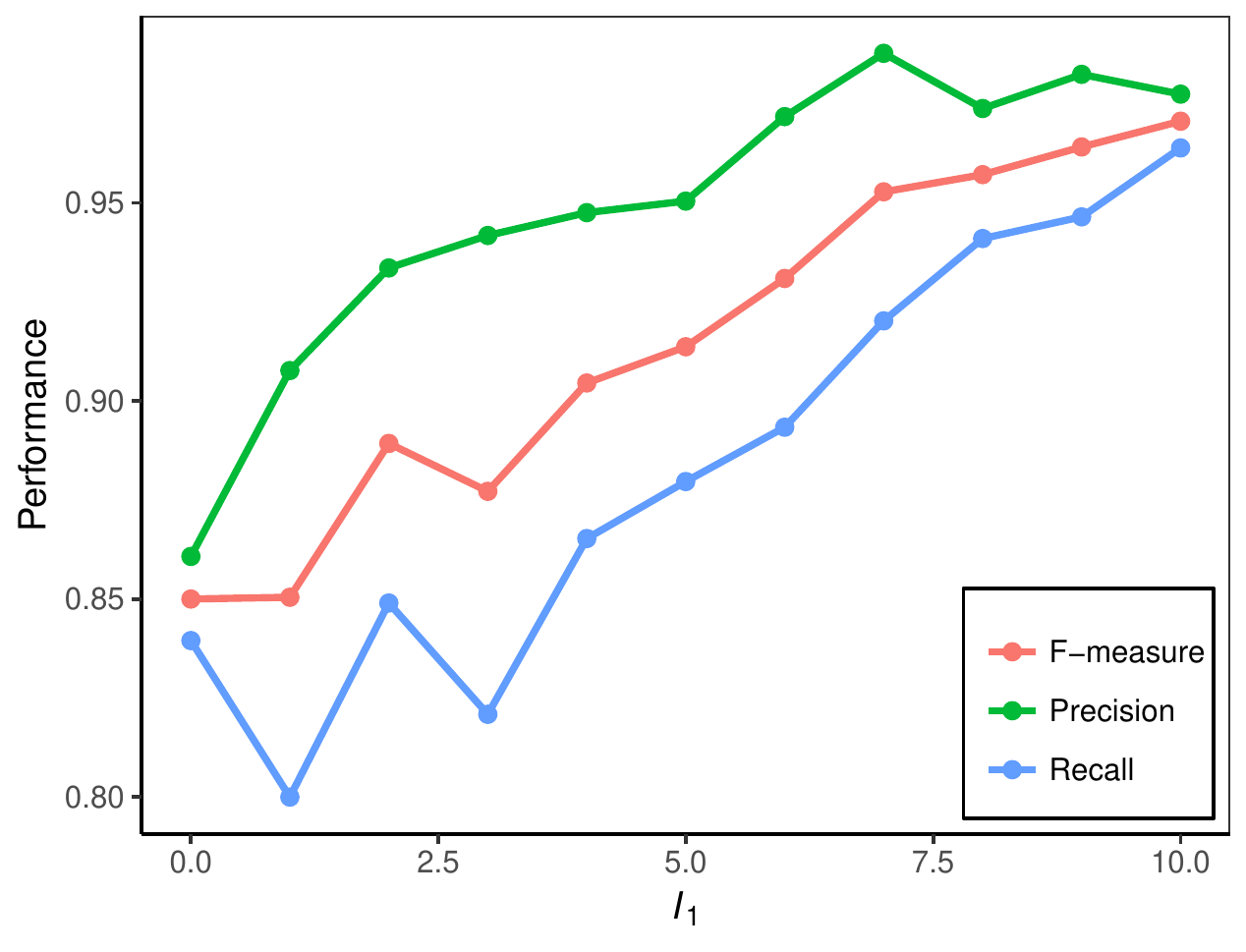}
\caption{Precision, Recall, and F-measure of the labels given by local inference vs. global inference when the value of $I_1$ changes.}
\label{fig7}
\vspace{-0.5cm}
\end{center}
\end{figure}

In Figure \ref{fig8}, we report the accuracy of the labels given by the global and local inference evaluated w.r.t. the true labels, as well as the accuracy of the local inference evaluated w.r.t. the labels provided by global inference. We find that as $I_1$ increases, both the accuracies of global and local inference w.r.t. the true labels increase significantly. In addition, 
the local inference 
gives similar result as the global inference (the green curve is high) and the accuracy increases as $I_1$ increases as well.
We also report in  Figure \ref{fig7} the precision, recall, and F-measure 
when comparing local inference with global inference, by treating label $1$ as positives. Both figures show that when $I_2$ is fixed and $I_1$ increases, which means that the correlation decay is stronger, our local inference method achieves better results.

\vspace{-0.3cm}

\section{Conclusion}
In this paper, we address query-specific marginal inference in large-scale graphical models using a new localized inference algorithm.  
We leverage the Dobrushin's comparison theorem to derive two error bounds for localized inference, 
including a simple bound based on graph distance and a localized bound from which we derive an efficient greedy expansion algorithm for constructing local regions for localized inference. 
Our experiments have shown that our bounds are practically useful and the algorithm works efficiently on various graphical models.  
Future directions include theoretical investigation on tighter bounds and development of more efficient greedy expansion algorithms. 
%

\bibliographystyle{apalike}
\bibliography{ref}

\clearpage

\section*{Appendix}
\subsection*{Proof of Theorem 2}
 \begin{proof}
 Let $m(i,j)=d(i,j)\ln\frac{1+(t-1)c}{tc}$, where $d(i,j)$ represents the distance between node $i$ and node $j$ in the graph and $t>1$. Then 
 \begin{align*}
& \max_{i\in I}\sum_{j\in I} e^{m(i,j)}C_{ij} \\
 = &\frac{1+(t-1)c}{tc}\max_{i\in I}\sum_{j\in I} e^{d(i,j)}C_{ij}\\
   \leqslant &\frac{1+(t-1)c}{t} \\
  < &1.
 \end{align*}
 Applying Lemma \ref{lem1}, we have
 \begin{align*}
 & \max_{k=\{1,-1\}}|\mu(x_i=k)-\nu(x_i=k)|\\
 \leqslant & \frac{e^{-m(i,\partial\alpha)}}{1-\frac{1+(t-1)c}{t}} \\
 = & \frac{e^{-d(i,\partial\alpha)\ln\frac{1+(t-1)c}{tc}}}{\frac{(t-1)(1-c)}{t}}.
 \end{align*}
 Substituting the inequality of $d(i,\partial\alpha)$ in the condition into right-hand side yields to the result. 
 \end{proof}
 
\subsection*{Proof of Corollary 2}
\begin{proof}
Note that the Dobrushin's interaction matrix $C$ of $\mu$ is a block diagonal matrix. Since there are no edges between $\alpha$ and $\beta$, the corresponding blocks equal to zero. If the Dobrushin condition holds, $D$ would also be a block-diagonal matrix and can be calculated easily from $C$. To see this, we have 
\begin{gather*}
\begin{matrix}
C=
\begin{blockarray}{ccc}
\alpha            & \beta \\
\begin{block}{[cc]c}
C_{\alpha\alpha}  & O & \alpha \\
O     			& C_{\beta\beta}& \beta \\
\end{block}
\end{blockarray} 
\end{matrix}\\
\begin{matrix}
\text{and} \quad D=\begin{blockarray}{ccc}
\alpha            & \beta \\
\begin{block}{[cc]c}
(I-C_{\alpha\alpha})^{-1}  & O & \alpha \\
O     			& (I-C_{\beta\beta})^{-1}  & \beta \\
\end{block}
\end{blockarray} \end{matrix}.
\end{gather*}
Applying the bound in Corollary \eqref{cor1} gives the result. 
\end{proof}

\subsection*{Calculation of $C$}
In the greedy approximation, suppose the true probability measure is $$\nu(\bm{x};\bm{\theta})=\frac{1}{Z_{\nu}(\theta)} \exp\big({{\sum_{\langle i\, j\rangle \in E}J_{ij}x_ix_j+\sum_ih_ix_i}\big)}.$$

We have approximate probability measure $$\mu(\bm{x};\bm{\theta})=\frac{1}{Z_{\mu}(\theta)} \exp\big({{\sum_{\langle i\, j\rangle \in E}\tilde{J}_{ij}x_ix_j+\sum_i\tilde{h}_ix_i}\big)}.$$

The coefficient $\tilde{J}_{ij}$ in $\alpha$ and $\beta$ are the same as in true probability measure $\nu$. Also, the coefficient $\tilde{h}_j$ in $\mathring{\alpha}$ and $\mathring{\beta}$ are the same as original graph. The difference is that $\tilde{J}_{ij}$ between boundaries becomes 0 and the coefficient $\tilde{h}_j$ in the boundaries has changed. We will calculate $C$ and $D$ as follows.

For the matrix of $D$, we need to get $C$ first. From the definition of $C_{ik}$, we have
\begin{equation*}
\frac{\mu_k(1|x)}{\mu_k(-1|x)}=\exp\left(2\sum_{j:d(j,k)=1}\tilde{J}_{jk}x_{j}+2\tilde{h}_k\right) \text{ and }
\end{equation*}
\begin{equation*}
 \mu_k(1|x)+\mu_k(-1|x)=1,
 \end{equation*}
where $d(j,k)$ stands for the distance between two nodes in the graph. Thus, we have
$$\mu_k(-1|x)=\frac{1}{\exp\left(2\sum_{j:d(j,k)=1}\tilde{J}_{jk}x_{j}+2\tilde{h}_k\right)+1}.$$

Similarly, $$\mu_k(-1|y)=\frac{1}{\exp\left(2\sum_{j:d(j,k)=1}\tilde{J}_{jk}y_{j}+2\tilde{h}_k\right)+1}.$$

Noting $|\mu_k(1|x)-\mu_k(1|y)|=|\mu_k(-1|x)-\mu_k(-1|y)|$, we have $C_{ik}=\sup\{\mu_k(-1|x)-\mu_k(-1|y):x=y \;\; \text{off} \;\;i\}$. If $d(i,k)\neq 1$, from the formula, we have $C_{ik}=0$. If $d(i,k)=1$, since $x=y \;\; \text{off} \;\;i$, without lose of generality, we assume $x_i=1$ and $y_i=-1$. Also, let $M=2\sum_{j:d(j,k)=1}\tilde{J}_{jk}x_{j}-2\tilde{J}_{ik}x_i+2\tilde{h}_k$. Then we have 
\begin{equation*}
\mu_k(-1|x)=\frac{1}{\exp(M+2J_{ik})+1},
\end{equation*}
\begin{equation*}
\mu_k(-1|y)=\frac{1}{\exp(M-2J_{ik})+1}.
\end{equation*}
By derivation, we know $|\mu_k(-1|x)-\mu_k(-1|y)|$ increases as $M$ increases from $-\infty$ to 0 while decreases when $M$ increases from 0 to $\infty$. Choosing $x$ to get optimal $M$ can be converted to a knapsack problem, which may be solved by dynamic programming or searching algorithm.

\subsection*{Calculation of $D$}
From the result above, we notice that if $i\in\alpha, k\in\beta$ or $i\in\beta, k\in\alpha$, then $C_{ik}=0$ because in approximation $\mu$, there is no edge between them. Hence, matrix $C$ is a block diagonal matrix. Suppose $C_{\alpha\alpha}$ and $D_{\alpha\alpha}$ stand for the matrix about nodes in $\alpha$, from linear algebra, we have $D_{\alpha\alpha}=(I-C_{\alpha\alpha})^{-1}$. In order to calculate the bound, we only need information about $D_{\alpha\alpha}$, thus $C_{\alpha\alpha}$. What's more, $\forall \;k\in\mathring{\alpha}, \forall \;i$, $C_{ik}$ would not change due to approximate. So, we can use some former results to reduce the calculation of $C_{\alpha\alpha}$. Meanwhile, when computing the inverse of $I-C$, we can also use former results by the property of block matrix as follows

\begin{tiny}
\begin{equation*}      
\left(                 
  \begin{matrix} 
    A_{11} & A_{12}\\  
    A_{21} & A_{22}\\
  \end{matrix}
\right)^{-1}
\end{equation*}
\begin{equation*}
=\left(
	\begin{matrix}
	(A_{11}-E)^{-1} & -(A_{11}-E)^{-1}A_{12}A_{22}^{-1}\\
	-A_{22}^{-1}A_{21}(A_{11}-E)^{-1} & A_{22}^{-1}+A_{22}^{-1}A_{21}(A_{11}-E)^{-1}A_{12}A_{22}^{-1}\\
	\end{matrix}
\right),
\end{equation*}
\end{tiny}

where $E=A_{12}A_{22}^{-1}A_{21}$. When we calculate the inverse, we can regard $I-C_{\partial\alpha\partial\alpha}$ as $A_{11}$, $I-C_{\mathring{\alpha}\partial\alpha}$ as $A_{21}$, $I-C_{\partial\alpha\mathring{\alpha}}$ as $A_{12}$, and
$I-C_{\mathring{\alpha}\mathring{\alpha}}$ as $A_{22}$. As we will discuss next, we only need $D_{ji}$, where $j\in\partial\alpha$. That means we only need to calculate $-(A_{11}-E)^{-1}A_{12}A_{22}^{-1}$.



\end{document}